\documentclass{IOS-Book-Article}

\usepackage{mathptmx}
\usepackage{soul}\setuldepth{article}
\def\hb{\hbox to 11.5 cm{}}

\usepackage{booktabs}
\usepackage{graphicx}
\usepackage{rotating}

\usepackage[numbers]{natbib}

\usepackage{tabularx}
\usepackage{dcolumn}
\newcolumntype{d}[1]{D{.}{.}{#1}}

\makeatletter
\newcolumntype{B}[3]{>{\boldmath\DC@{#1}{#2}{#3}}c<{\DC@end}}

\usepackage{subcaption}
\usepackage[gen]{eurosym}

\begin{document}
\pagestyle{headings}
\def\thepage{}
\begin{frontmatter}              

\title{Investigating Labeler Bias in Face Annotation for Machine Learning}

\author[A,B]{\fnms{Luke} \snm{Haliburton}\orcid{0000-0002-5654-2453}\thanks{Corresponding Author: Luke Haliburton, luke.haliburton@ifi.lmu.de.}},
\author[A]{\fnms{Sinksar} \snm{Ghebremedhin}\orcid{0000-0002-2874-2909}},
\author[C]{\fnms{Robin} \snm{Welsch}\orcid{0000-0002-7255-7890}},
\author[A]{\fnms{Albrecht} \snm{Schmidt}\orcid{0000-0003-3890-1990}},
and
\author[A]{\fnms{Sven} \snm{Mayer}\orcid{0000-0001-5462-8782}}

\address[A]{LMU Munich, Germany}
\address[B]{Munich Center for Machine Learning (MCML), Germany}
\address[C]{Aalto University, Finland}

\begin{abstract}
In a world increasingly reliant on artificial intelligence, it is more important than ever to consider the ethical implications of artificial intelligence. One key under-explored challenge is labeler bias --- bias introduced by individuals who label datasets --- which can create inherently biased datasets for training and subsequently lead to inaccurate or unfair decisions in healthcare, employment, education, and law enforcement. Hence, we conducted a study ($N$=98) to investigate and measure the existence of labeler bias using images of people from different ethnicities and sexes in a labeling task. Our results show that participants hold stereotypes that influence their decision-making process and that labeler demographics impact assigned labels. We also discuss how labeler bias influences datasets and, subsequently, the models trained on them. Overall, a high degree of transparency must be maintained throughout the entire artificial intelligence training process to identify and correct biases in the data as early as possible.
\end{abstract}

\end{frontmatter}
\markboth{\hb}{\hb}

\section{Introduction}
\label{section:intro}
Artificial intelligence (AI) is rapidly becoming involved in numerous areas of life, making far-reaching decisions such as granting loans and hiring people. Amazon analyzes customers' purchasing behavior\footnote{\href{https://www.gigaspaces.com/blog/amazon-found-every-100ms-of-latency-cost-them-1-in-sales}{https://www.gigaspaces.com/blog/amazon-found-every-100ms-of-latency-cost-them-1-in-sales}}, 
Netflix studies entertainment preferences\footnote{\href{https:/about.netflix.com/en/news/four-years-after-house-of-cards-netflix-members-elect-their-owntv-schedule}{https:/about.netflix.com/en/news/four-years-after-house-of-cards-netflix-members-elect-their-owntv-schedule}}, 
and Facebook uses social interactions to tailor content to their users~\cite{horwitz2021facebook}. Data collection, processing, and prediction are key pillars of AI applications. Although AI is a powerful tool, the fundamental reliance on data can be problematic due to the potential for bias to be embedded in datasets, creating unintended consequences. One under-investigated contributing factor to biased AI tools is labeler bias, which results from cognitive biases~\cite{eickhoff2018cognitive} in crowd workers and other dynamics in the labeling process~\cite{miceli_between_2020}. Many AI applications rely on crowdsourcing platforms to label their data, yet they usually do not consider whether they are utilizing a diverse population of labelers~\cite{miceli_studying_2022}. A biased labeler pool could lead to unfair outcomes for certain groups, such as women, ethnic minorities, or people from disadvantaged neighborhoods. Therefore, it is crucial to examine labeler pools with a critical lens to avoid bias and create a more fair and transparent process.

Investigating labeler bias is essential to understand how labelers influence datasets, but existing studies in this area are limited in scope. Recent work has demonstrated that rater identity plays a significant role in labeling toxicity for online comments~\cite{goyal_is_2022,sap_annotators_2022}. One critical paper by Bender et al. \cite{bender2021dangers} sheds light on how human biases can be unintentionally perpetuated in machine learning, highlighting that biases introduced in the labeling stage can propagate through to end decisions. In response to this issue, researchers in machine learning have attempted to model and correct for bias effects~\cite{jiang_identifying_2020,wauthier_bayesian_2011,cohn_modelling_2013}. In general, bias can be partially attributed to stereotypes, which occur when one assigns traits to an individual based on preconceived notions about their group~\cite{fiske2018stereotype}. The Stereotype Content Model (SCM) is an established practical theory explaining stereotypes, such as perceived warmth and competence, that has been applied in Human-Computer Interaction (HCI) (e.g., \cite{schwind2019understanding,marsden_stereotypes_2016}). However, there is a lack of work applying the SCM to characterize biases introduced into datasets by crowdsourced labelers.

In this paper, we address the gap in the existing literature by investigating stereotypes and bias in labeling tasks. We conducted a survey ($N$=98) asking crowd workers to label a series of human faces from the FairFace dataset~\cite{karkkainen2021fairface}. We selected faces with equal representation from seven ethnicities and two sexes and recruited labelers with the same balanced demographic distribution. We asked labelers to rate the portraits based on income and perceived warmth, competence, status, and competition. In this way, we investigate relationships between stereotype perceptions and income within and between demographic groups. We developed this study design to answer the following research questions:

\begin{itemize}
    \item[\textbf{RQ1}] What is the impact of \textit{stereotypes held by labelers} on their estimates of others in labeling tasks?
    \item[\textbf{RQ2}] What is the impact of the \textit{ethnicity or sex of labelers} on their estimates of others in labeling tasks?
\end{itemize}

Our results show that labelers possess stereotypes independent of their own demographics and that these stereotypes impact the labels they assign. Further, we show that the ethnicities of both labelers and portraits impact the predictions assigned by the labelers\footnote{Please note from the outset that we do not analyze \textit{which} groups are biased in which directions. Our aim is not to apply judgments to specific groups but rather to investigate whether ratings are impacted by demographics in an effort to promote fairness and ethics in machine learning.}. Our results indicate that labeler bias is a function of both labeler demographics and characteristics of the labeled subject, suggesting that recruiting a diverse labeler pool may not be enough to counteract the bias. Overall, this paper contributes evidence for the existence of labeler bias and discusses its consequences. In particular, we contribute an investigation of bias in the context of a face-labeling task using a publicly available dataset. Our findings raise awareness for labeler bias, which we hope leads to researchers and practitioners critically examining and revisiting current practices in data labeling.

\section{Related Work}
In this section, we first present prior work on data bias in machine learning, examining historical bias and labeler bias. Next, we introduce the stereotype content model (SCM) and how it relates to computer science and our work.

\subsection{Bias in Machine Learning}
Biases can be introduced in machine learning models and algorithms at multiple stages. Algorithms can contain systematic biases embedded by the moral concepts of developers~\cite{hagendorff2020ethics,tsamados2021ethics}. However, data bias is a more prevalent concern in intelligent systems. As Mueller \cite{muller2021ethics} describes, the quality of a system is coupled with ``the quality of the data provided, following the old slogan \textit{garbage in, garbage out}.'' It follows that an AI system will make biased decisions if it is trained on biased data. Notable sources of bias, which we detail below, include historical bias, non-representative sampling, and labeler bias.

\subsubsection{Historical and Sampling Bias}
Historical bias occurs when a system is trained on data resulting from real, biased scenarios. In an early example, St. George's University Medical School introduced a new computer system that systematically denied admission to women and people with `foreign-sounding names' based on historical data~\cite{schwartz2019untold,lee2019algorithmic}. Similar issues persist in many modern systems. A CV screening tool at Amazon preferentially hired men even after explicit references to gender, race, and sexual orientation were removed from the dataset~\cite{kodiyan2019overview}.
Algorithms for crime prediction typically rely on historical crime data in which ethnic minorities are over-represented\footnote{\href{https://blogs.scientificamerican.com/voices/i-know-some-algorithms-are-biased-because-i-created-one/}{https://blogs.scientificamerican.com/voices/i-know-some-algorithms-are-biased-because-i-created-one/}}. 
Even using online proxies to remove sensitive characteristics continues to lead to biased decisions~\cite{lee2019algorithmic,zarsky2015understanding}.  Such systems can have a drastic impact on the lives of real people.  For example, minority groups have an increased likelihood of being stopped and searched by the police, based only on immutable characteristics which they cannot control~\cite{angwin2016machine}.

Generating datasets based on emerging data can also lead to biased results. Bender et al. \cite{bender2021dangers} describe the experience of collecting data through the internet: ``white supremacist and misogynistic, ageist, etc. views are over-represented in the training data, not only exceeding their prevalence in the general population but also setting up models trained on these datasets to amplify biases and harms further.'' Women are also underrepresented on platforms used for data collection~\cite{barera2020mind}, which results in decisions based on male-skewed data sources. Although issues of historical bias and non-representative sampling continue to be an issue, they are not the focus of this paper. Rather, we aim to investigate the under-explored potential impact of bias within the people labeling data to generate biased datasets.

Sampling bias has also been identified in face annotation tasks. Da Silva and Pedrini \cite{silva_effects_2015} found that an emotion classifier trained on one cultural group was inaccurate when used on a different cultural group. Scheuerman et al. \cite{scheuerman_how_2020} recommend embedding race and gender information into databases and classifications to increase transparency.

\subsubsection{Labeler Bias}
Labeler bias occurs when individuals annotate a dataset and embed bias into the resulting data. This often occurs unintentionally, as Wall et al. \cite{wall2018four} have argued that unconscious biases can influence judgments and lead to inaccurate conclusions in visual analytic feedback tasks. In one poignant example of labeler bias, prior work has shown that people from Western cultures tend to rate people from other cultures as being less attractive than themselves~\cite{hutson2018debiasing}. Consequently, an attractiveness dataset labeled primarily by people from Western cultures risks having this bias embedded\footnote{Attractiveness Test: \href{https://attractivenesstest.com/}{https://attractivenesstest.com/}}. Any system that uses such a dataset as a basis for decisions would subsequently perpetuate the bias of the labelers. Past work in machine learning has shown that fairness and accuracy can be aligned, motivating a push towards more fairness in training as a simultaneous push towards increased accuracy~\cite{wick_unlocking_2019}. Prior research has found that even highly experienced labelers fail to produce unbiased labels~\cite{hube_understanding_2019}. 

Perhaps most relevant to our study is recent work in CSCW by Goyal et al. \cite{goyal_is_2022} on rater identity. They found that rater identity (i.e., African American, LGBTQ, or neither) significantly influenced how raters annotated toxicity in online comments. They suggest that raters who self-identify with the identities targeted in online comments provide additional nuance and more inclusivity in trained models. Similar work found that other social variables (e.g., conservatism) also impact toxicity labels~\cite{sap_annotators_2022}. Past work has identified that labeler bias also stems from socio-economic contexts and the application of power structures within annotation companies~\cite{miceli_between_2020}. In response, researchers have proposed more fair and human-centric crowdsourcing frameworks considering demographics and appropriate compensation~\cite{barbosa2019rehumanized}. In the presented examples, we can see that labeler bias can stem from a number of characteristics of the labeling group. In this paper, we extend the body of work on labeler bias by exploring whether demographic clusters of labelers can have an impact on labeling task estimates.

Past work has attempted to find solutions to account for labeler bias. Several strategies to model labeler bias include using knowledge about the ground truth~\cite{jiang_identifying_2020}, bayesian methods~\cite{wauthier_bayesian_2011}, or multi-task Gaussian Processes~\cite{cohn_modelling_2013}. Geva et al. \cite{geva_are_2019} recommend that labelers for testing and training datasets be distinct groups since they found that subjective NLP labels produced by a group do not generalize well. Instructions for annotators have also been found to embed bias~\cite{parmar_dont_2022,miceli_data-production_2022}. In the context of face annotation, Engelmann et al. \cite{engelmann_what_2022} argue that `secondary' (i.e., subjective) characteristics may not be appropriate attributes for facial recognition systems to predict.

\subsection{Stereotype Content Model}
The Stereotype Content Model (SCM) is a theory that explains how people develop stereotypes about other people. The SCM proposes that people primarily use two dimensions, warmth and competence, to assess other people. These dimensions prognosticate emotional prejudices, which in turn lead to discrimination~\cite{fiske2018stereotype,fiske2007universal,fiske2002model}. People group themselves based on what they perceive to be similarities between themselves and others. Across-group prejudice is a judgment on an emotional, cognitive, or behavioral level towards another group~\cite{dovidio2010intergroup}.

The warmth-competence model is a well-researched \cite{durante2017poor,grigoryev2019mapping,diamantopoulos2017explicit,schwind2019understanding,lin2005stereotype} fundamental theory in social psychology with broad implications for both social perception and social interaction~\cite{cuddy2008warmth}. The model helps to explain how we form stereotypes about different groups of people, positing that we judge groups based on how warm and competent we perceive them to be~\cite{fiske2015intergroup,cuddy2009stereotype}. Warmth is the evaluation criterion by which a person is perceived as friendly and trustworthy, while competence shows how capable and compelling a person is perceived to be. Generally, our reaction to individuals from certain groups is governed by how warm and competent we perceive them to be. For example, people generally see elderly individuals as warm but not competent and therefore react with pity. Groups seen as competent but not warm are met with envy, while those perceived as both competent and warm tend to be admired~\cite{fiske2002model}. The SCM has also been extended to include competition and status, which are particularly influenced by how an individual views a specific group relative to their own group~\cite{fiske2002model}.

Nicolas et al.~\cite{nicolas2021comprehensive} made a significant contribution by developing dictionaries for stereotypical content. These dictionaries simplify the study of stereotypes and speed up the identification of social biases in AI, social media, and other textual sources. Subsequent work has used these dictionaries to automatically identify the stereotypical language in news coverage~\cite{kroon2021guilty} or to mitigate stereotypical language through anti-stereotypes~\cite{fraser2021understanding}. The SCM has also been used to understand human-machine interactions. For example, McKee et al.~\cite{mckee2021understanding} used the SCM to explore how individuals react to different digital avatars. They found that users increasingly perceive the system as being warm if it appears in the role of an assistant and cold if it appears in the role of a competitor. The SCM has also been used in HCI to investigate stereotypes in personas~\cite{marsden_stereotypes_2016} and determine the social acceptability of mobile devices~\cite{schwind2019understanding}. In our work, we are investigating the way that labeler stereotypes influence the labels they assign during annotation tasks. As such, we use the SCM, a well-established model, to understand how stereotypes vary across our labeler population.

\section{Method}
We conducted an online survey where participants from various demographic backgrounds labeled portraits of varying ethnicities and sexes based on income and the SCM. For this, we balanced the participants' self-reported ethnicity.

Since ethnicity labels are not clearly defined~\cite{phinney1996when}, we aligned our ethnicity categorization with the FairFace dataset. The following seven groups listed in FairFace will be referred to as ethnicities\footnote{We use the term ``ethnicity,'' as it encompasses more social aspects and is a broader term than ``race,'' although they are often used interchangeably in practice. We also note that ethnic distinctions are non-specific, but such labels are commonly used in Machine Learning applications such as the FairFace dataset.} in this work: (1) Black, (2) East Asian, (3) Indian, (4) Latino, (5) Middle Eastern, (6) Southeast Asian, and (7) White.

\begin{figure*}[t]
 \centering
	\begin{subfigure}[t]{0.24\linewidth}
	    \includegraphics[width=\linewidth]{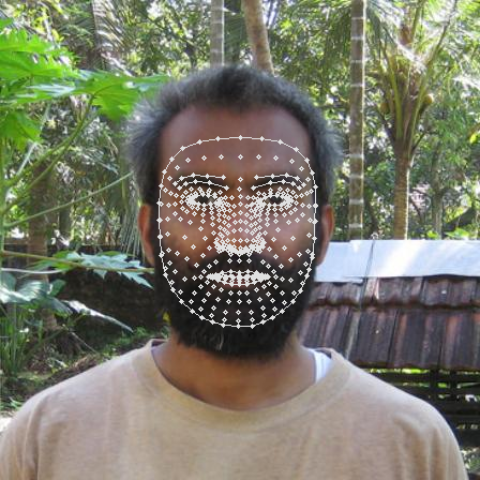}
	    \caption{Face is frontal, included.}
	\label{fig:image_positions:a}
	\end{subfigure}%
	\hfill
	\begin{subfigure}[t]{0.24\linewidth}
	    \includegraphics[width=\linewidth]{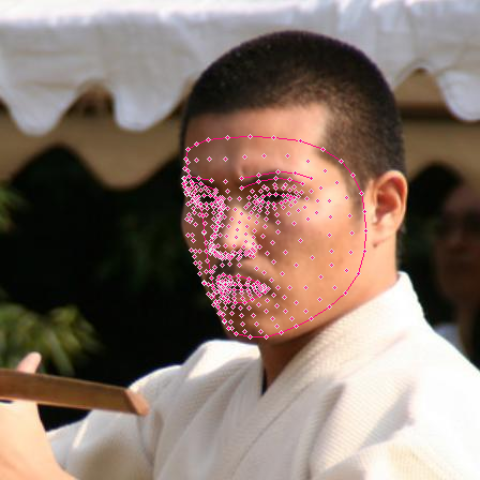}
	    \caption{Eyes are not centered, excluded.}
	\label{fig:image_positions:b}
	\end{subfigure}%
	\hfill
	\begin{subfigure}[t]{0.24\linewidth}
	    \includegraphics[width=\linewidth]{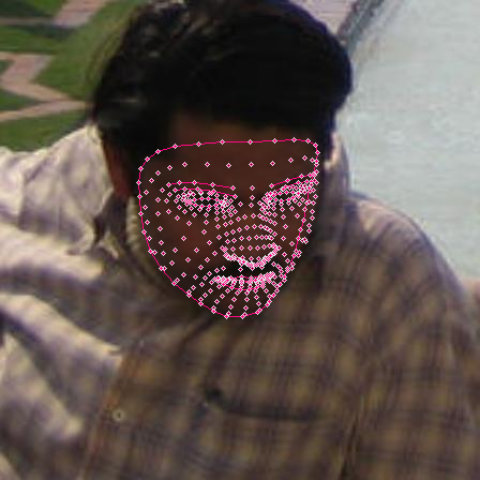}
	    \caption{Face is not centered on y-Axis, excluded.}
	\label{fig:image_positions:c}
	\end{subfigure}%
	\hfill
	\begin{subfigure}[t]{0.24\linewidth}
	    \includegraphics[width=\linewidth]{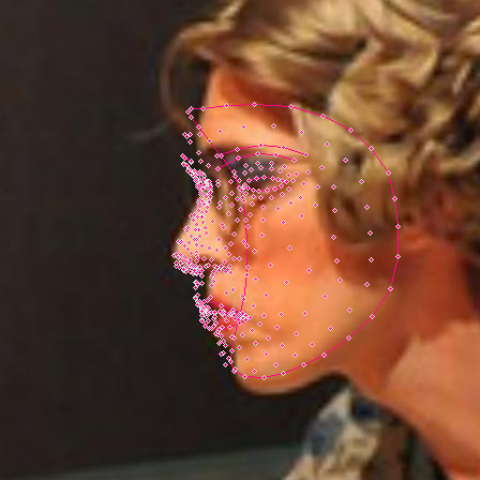}
	    \caption{Distance of edges is not centered, excluded.}
	\label{fig:image_positions:d}
	\end{subfigure}%
    \caption{We implemented a script to exclude non-frontal faces and images containing more than one person.}%
    \label{fig:image_positions}%
\end{figure*}

\subsection{Dataset Preprocessing and Portrait Selection}\label{section:preprocessing}
For our study, we selected portraits from the FairFace dataset (License CC BY 4.0)~\cite{karkkainen2021fairface}. This dataset consists of images of people and was specifically developed to be balanced in ethnicity, sex, and age. The authors generated this dataset to foster the development of fair and inclusive machine learning models. We selected the FairFace dataset because it provides us with a high probability of finding high-quality images across a wide range of demographic categories. The ethnicity, gender, and age tags for the images were labeled by Amazon Mechanical Turk (MTurk) users based on a two-thirds majority vote~\cite{karkkainen2021fairface}. The majority vote labeling process is common practice for labeling~\cite{zhang2016learning}, but consequently, we do not have a ground truth for the labels. However, to our knowledge, this is the most extensive dataset available with diverse ethnicities and sexes and was, therefore, the most suitable choice for this study.

We defined several criteria to filter the images in the dataset and create a subset for our study. Our research questions concern ethnicity and sex, so we designated age as a control variable and only selected images in the dataset within the range of 27 to 40 years old (the narrowest age filter provided by the dataset). To avoid any confounding factors, we also defined the following criteria:
1) Only one visible, camera-facing person with a neutral facial expression, 
2) A neutral background with no cropped edges, 
3) In color (i.e., no black \& white images), and
4) No glasses/sunglasses/headgear.

As the FairFace dataset contains more than 100,000 images, we could not manually filter all images by our predefined criteria. Therefore, we created a preprocessing script with the following functions: \textbf{(1)} We removed any images with age labels lower than 27 or higher than 40 \textbf{(2)} We detected faces within each image using the python library ``face-recognition''\footnote{\href{https://github.com/ageitgey/face_recognition}{https://github.com/ageitgey/face\_recognition}}
and removed any images where the number of faces was equal to zero or greater than one. \textbf{(3)} We detected face landmarks using the python library \textit{mediapipe}~\cite{lugaresi2019mediapipe} and used these landmarks (see \autoref{fig:image_positions}) to remove images where the subject is not facing the camera. A face is considered to be non-frontal if it deviates on the x-axis (see \autoref{fig:image_positions:b}) or the y-axis (see \autoref{fig:image_positions:c}) beyond a threshold of 0.09, or if both thresholds are crossed, indicating that the face is completely non-frontal ((see \autoref{fig:image_positions:d}). If none of the thresholds are crossed, the face is considered to be frontally aligned  (see \autoref{fig:image_positions:a}). \textbf{(4)} We detected facial expressions using a machine learning model based on\footnote{\href{https://github.com/priya-dwivedi/face\_and\_emotion\_detection}{https://github.com/priya-dwivedi/face\_and\_emotion\_detection}}. 
We selected only images with neutral facial expressions. \textbf{(5)} Finally, we were left with 1,834 images which we manually filtered. Three authors evaluated each of the remaining images and only selected those that fit all of the above criteria. The three authors triple-coded and only included images where all agreed. This resulted in 56 portraits, four for each sex and ethnicity combination.

\subsection{Participants} \label{section:participants}
\begin{figure*}[t]
    \centering
    \subfloat[\centering Participant income by ethnicity] {{\includegraphics[width=.49\textwidth]{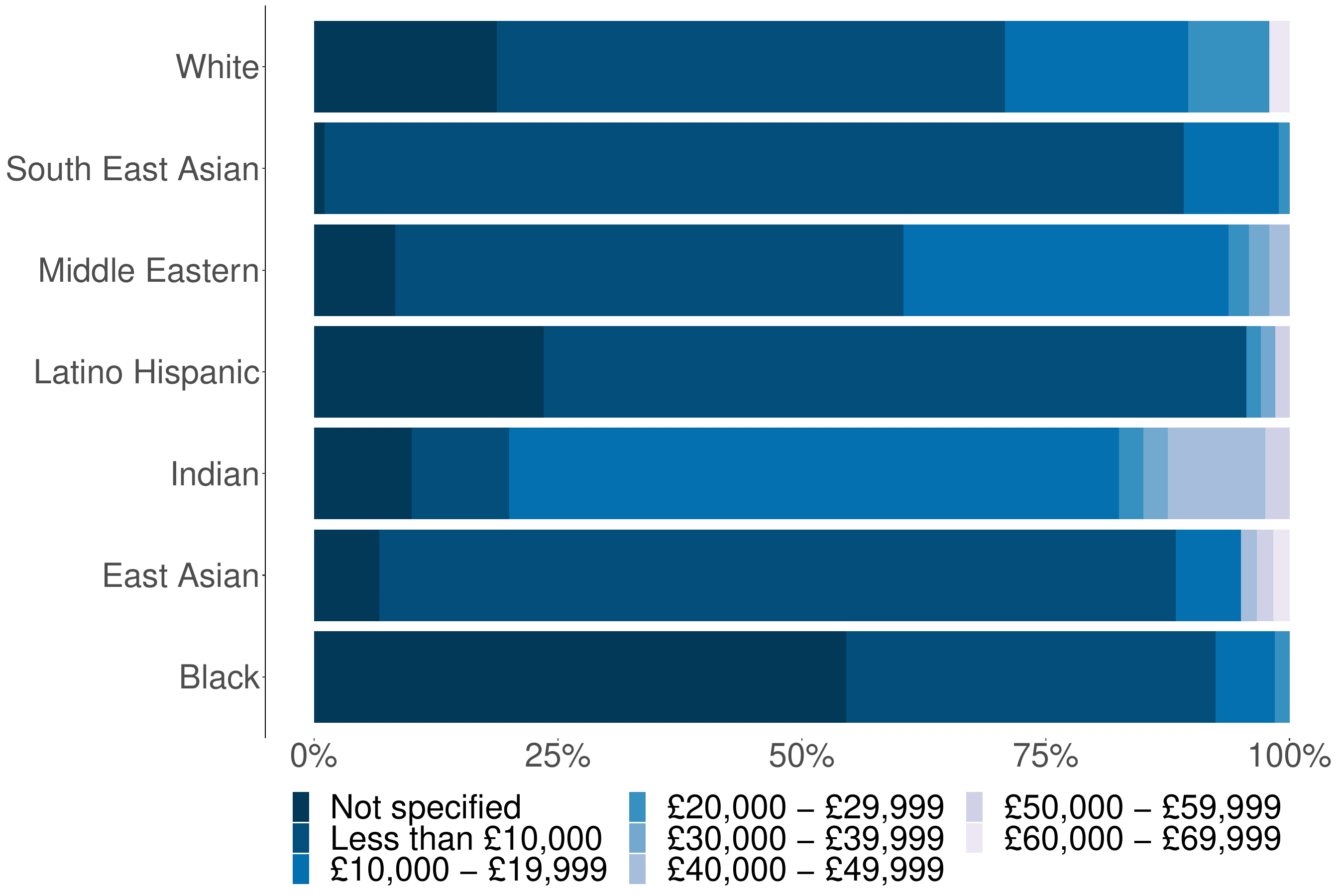} }}%
    \hfill
    \subfloat[\centering Participant income by sex] {{\includegraphics[width=.49\textwidth]{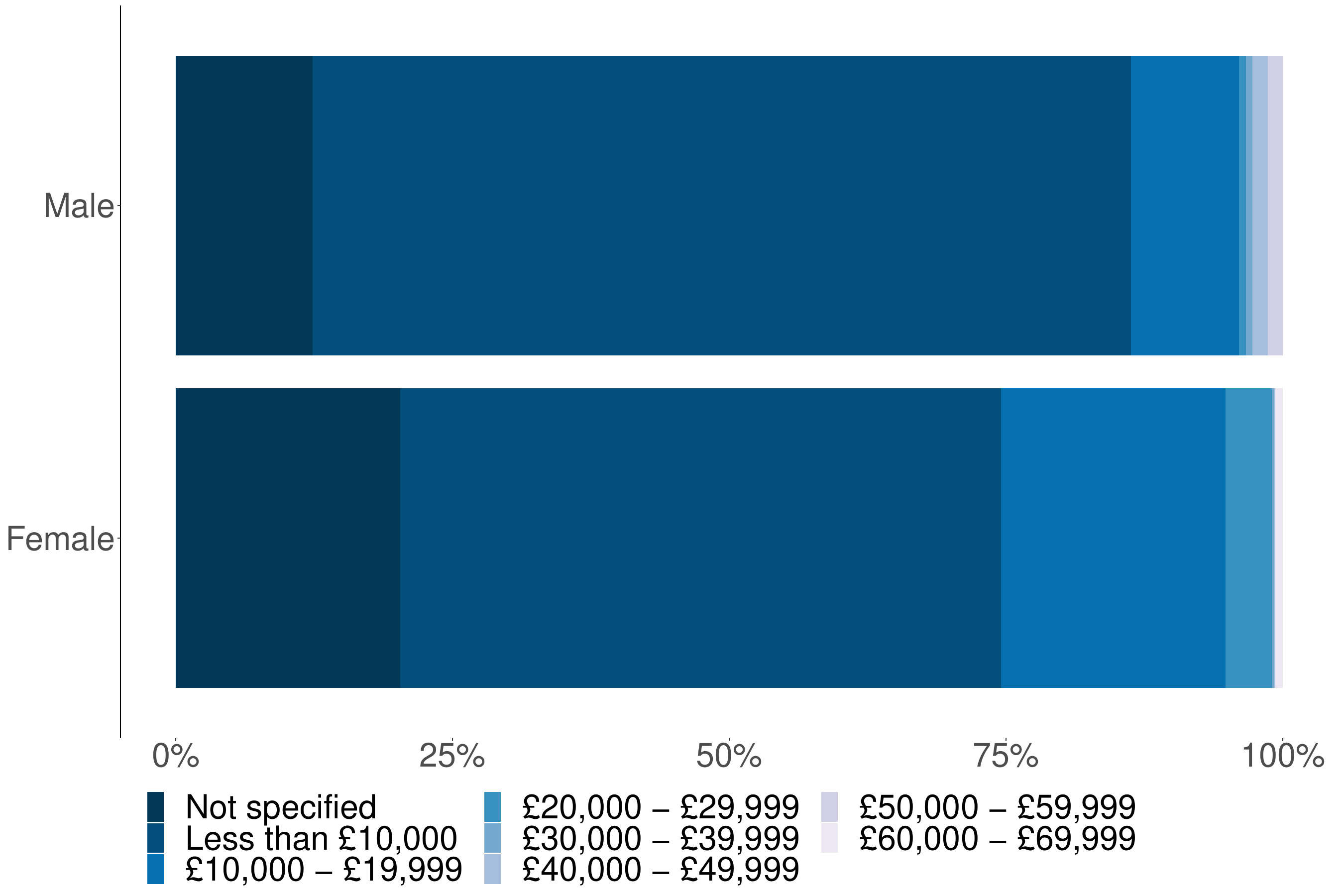} }}%
    \caption{The average participant income by ethnicity and age. A Pearson correlation showed that income variation had no significant impact on the results. Amounts are shown in GBP (\pounds{}) as this is the currency used by Prolific.}%
    \label{fig:labeler_income}%
\end{figure*}

We recruited 98 participants (49 female and 49 male) from Prolific\footnote{Prolific: \href{https://www.prolific.co}{https://www.prolific.co}}. Participants were between 18 and 52 years old ($M = 26.1$, $SD = 6.9$). The participants were equally distributed among the seven ethnicity categories from the FairFace dataset, and age. They live in 21 countries\footnote{Full study information is provided at \url{https://github.com/mimuc/labeler-bias}.\label{note:github}}. Participants were compensated at a rate of 10\euro{} per hour for a total of 3\euro{}. The study was approved by the ethics committee within the LMU Munich University Faculty. Participants' income, as indicated in \pounds{} on Prolific, varied across ethnicities, shown in \autoref{fig:labeler_income}. We analyzed the correlation between participant income and their income estimates for the portraits to prevent estimation bias. A Pearson correlation showed no significant correlation ($r=-.025$, $p=.403$), so we assume that participants' own income does not influence the results.

\subsection{Study Procedure}
\label{section:procedure}
We used Prolific, a crowdsourcing marketplace, to gather data as it can provide demographic information about the participants. We created a separate posting for each sex and ethnicity category and used built-in demographic filtering features to ensure that we recruited an equal number of participants from each category.

The participants were first given a brief overview before providing informed consent and completing a demographic questionnaire. The demographic information was also provided by Prolific, but we collected it in the survey to ensure that the ethnicity and sex labels matched the terminology used in the dataset. We then asked participants to respond to questions associated with portraits of people. Each participant was presented with one randomly chosen portrait from each ethnicity and sex category, resulting in 14 portraits per participant. 

For each portrait, the participants responded to SCM questions about perceptions of warmth, competence, status, and competition based on~\cite{fiske2002model}\textsuperscript{\ref{note:github}}. Several sub-scale questions are averaged to score ratings of warmth, competence, status, and competition on a scale. The SCM is an established method of measuring stereotype attitudes~\cite{fiske2002model}. We also asked participants to estimate the income for each portrait. To prevent country- and currency-specific biases, participants were asked to assign a value between ``high'' or ``low'' with a slider rather than a dollar amount. This task mimics several realistic scenarios. For example, it is customary in some European countries to include a photo on a CV when applying for a job. Hiring personnel, therefore, make judgments associating income and suitability for a job based on a photo of a face. Finally, we included attention checks (correctly answering a multiple choice question about information written in a short text) in the survey to prevent spam responses, which is common practice in crowdsourced tasks~\cite{abbey2017attention}.


\section{Results}
To investigate the relationship between demographics and labels, we performed two-way ANOVA models (Type III, $\alpha$ = .05) using Greenhouse-Geisser correction~\cite{greenhouse1959methods} where the sphericity assumption is violated. Note that although it would be possible to compare all levels on all factors and their interaction with post hoc tests, we refrain from doing so. First, test-corrections will be very conservative for pairwise comparisons due to the high number of levels. Second, we were only interested in showing that our factors can explain variation on the dependent variable. As such, we do not analyze which specific biases are present in specific groups.


\begin{figure*}[t]
 \centering
  \begin{subfigure}[b]{0.48\linewidth}
 \includegraphics[width=\linewidth]{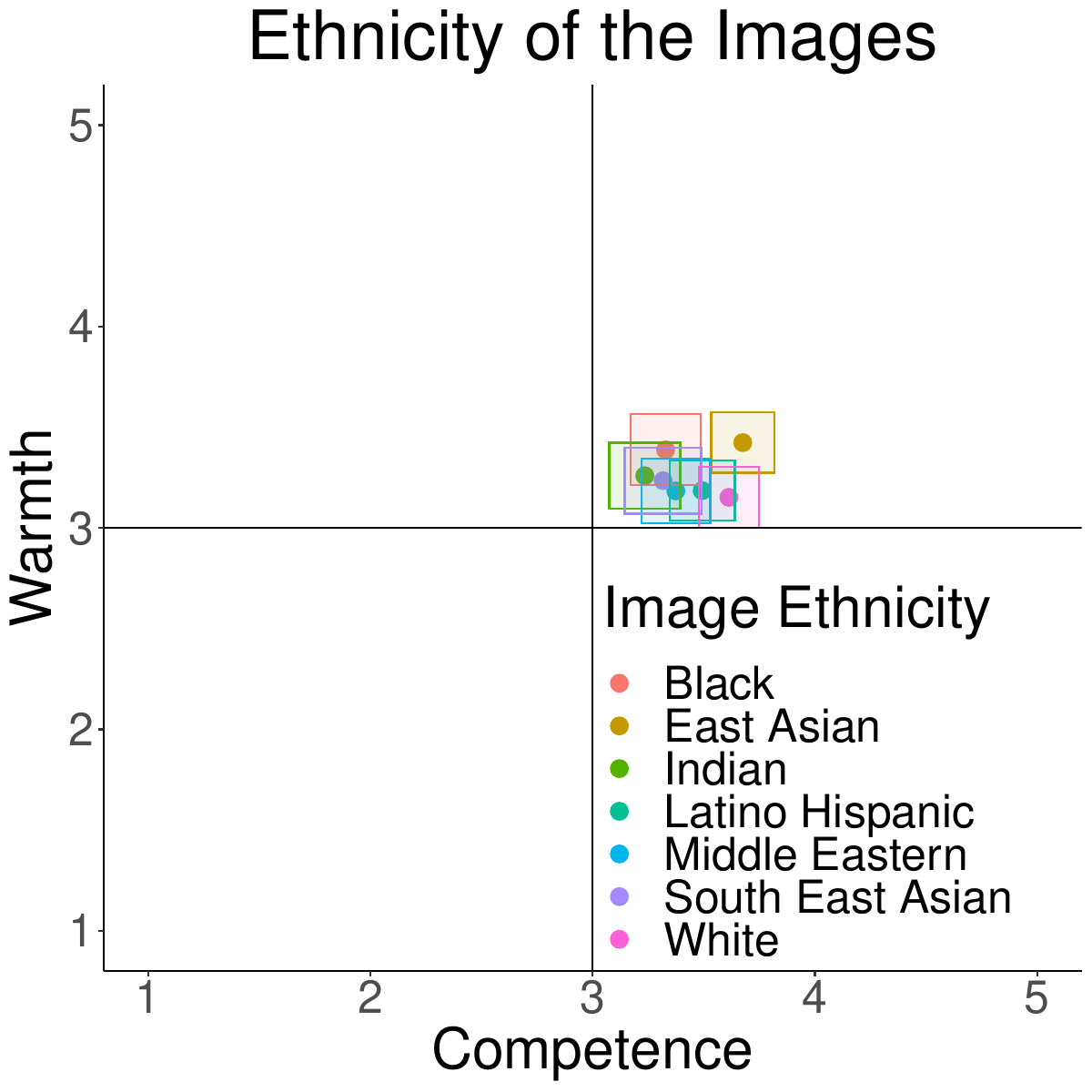}
 \subcaption{\textsc{Portraits}}
 \label{fig:scm_portraits}
 \end{subfigure}
 \hfill
 \begin{subfigure}[b]{0.48\linewidth}
 \includegraphics[width=\linewidth]{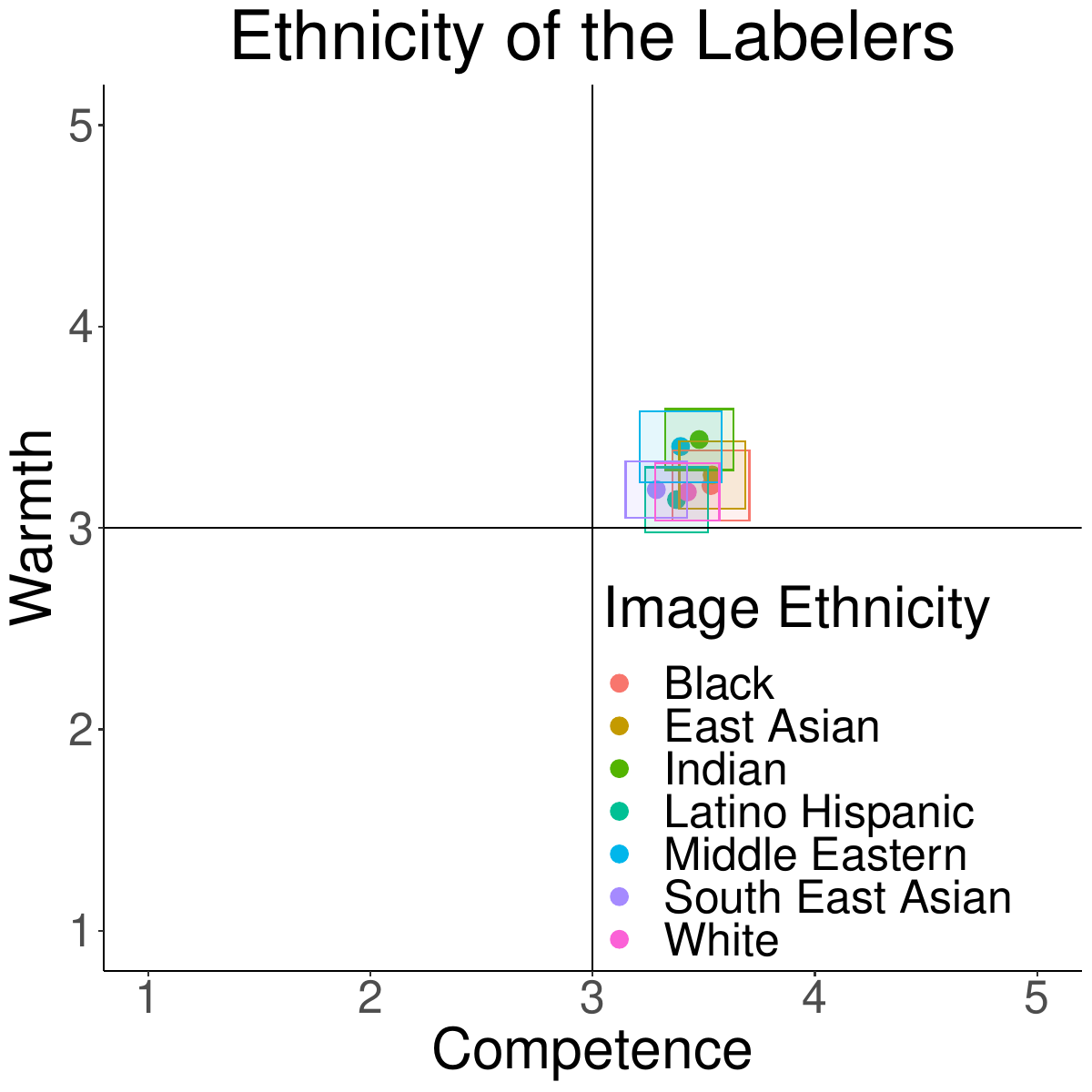}
 \subcaption{\textsc{Labelers}}
 \label{fig:scm_labelers}
 \end{subfigure}
 \caption{Warmth-Competence ratings displayed from the perspective of the portraits and the labelers, including a 95\% confidence interval. All ratings are clustered near neutral (3) for both warmth and competence.}
 \label{fig:scm_ratings}
\end{figure*}

\subsection{The Impact of Stereotypes on Estimations (RQ1)}
In line with the SCM~\cite{fiske2002model}, \autoref{fig:scm_ratings} shows the Warmth-Competence ratings assigned to the portraits by the labelers. \autoref{fig:scm_portraits} shows the stereotypes assigned to each \textsc{Portrait$_{Ethnicity}$} while \autoref{fig:scm_labelers} presents the stereotypes assigned by each \textsc{Labeler$_{Ethnicity}$}. All ratings are clustered near neutral warmth and neutral competence.

We conducted a Pearson correlation analysis for warmth, competence, status, and competition against the estimated income. \autoref{tab:pearson} shows that three out of four stereotype variables have a significant positive correlation with income. Competence, status, and competition all covary significantly with income, with status being the most positively correlated. Only warmth did not significantly covary with the income estimates.


\begin{table*}[t]
\caption{The Pearson correlations for each of the stereotype variables and the estimated income. We also computed Linear mixed models that take into account the nested structure in the data. However, showed no noteworthy difference from the simple correlations, so we only report the Pearson correlations for brevity.}
\label{tab:pearson}
\begin{center}
\begin{tabularx}{.4\textwidth}{Xd{3.3}d{3.3}}
\toprule
 &  \multicolumn{1}{c}{\textit{p}} & \multicolumn{1}{c}{\textit{r}} \\ 
\midrule
Warmth         &   .093          & .045 \\
Competence     & < .001          & .541 \\
Status         & < .001          & .773 \\
Competition    & < .001          & .431 \\
\bottomrule
\end{tabularx}
\end{center}
\end{table*}



\begin{figure*}[t]
 \centering
 \includegraphics[width=\textwidth]{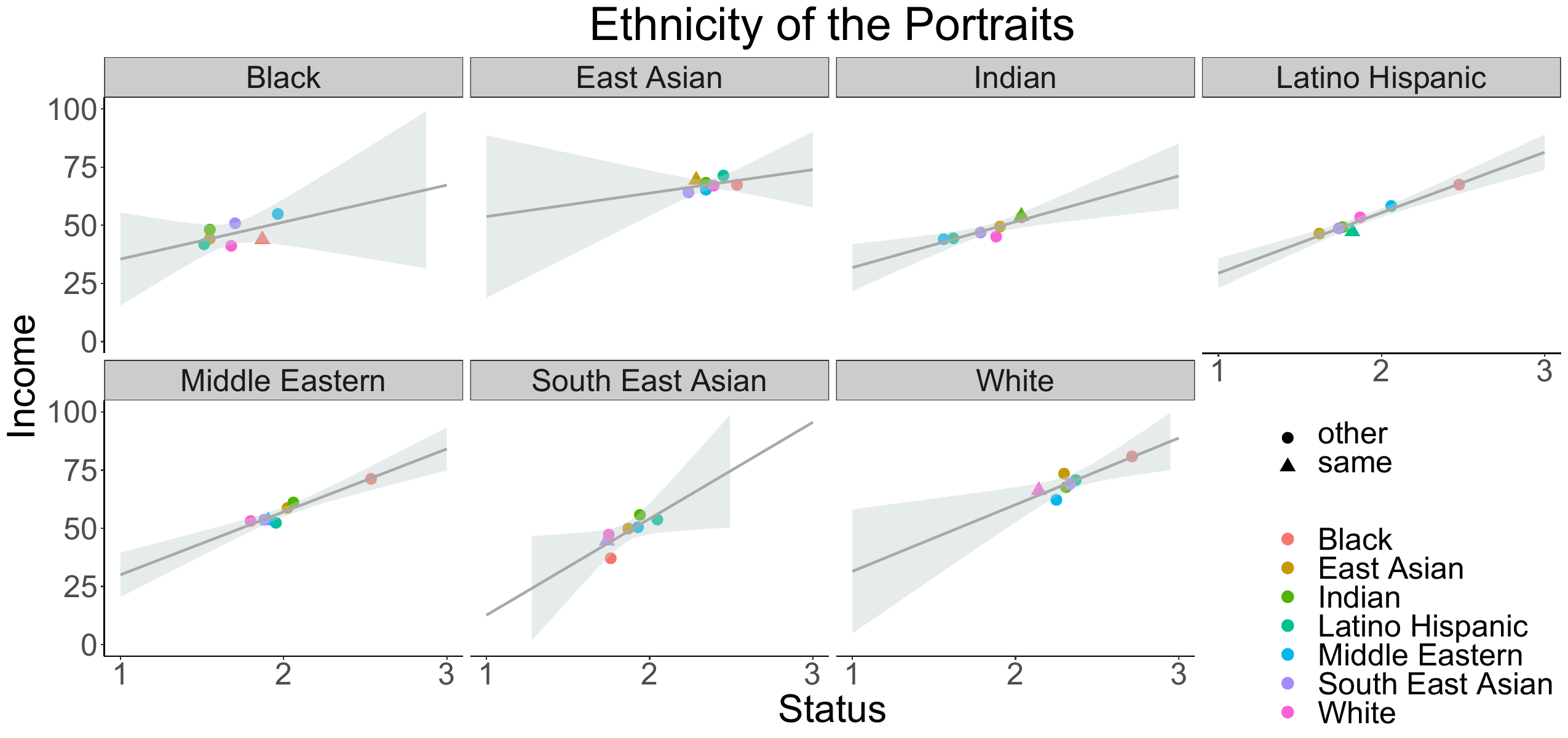}
 \caption{Correlation between mean status and income. Each subplot represents a portrait ethnicity and the points in each plot show how labelers of each ethnicity rated the portraits.}
 \label{fig:meanIncome_X_sumFactorStatus}
\end{figure*}

In general, the biases were in the predicted directions~\cite{fiske2002model}. For example, participants generally estimated a low income for a person they rated as low status, and vice versa. \autoref{fig:meanIncome_X_sumFactorStatus} illustrates this phenomenon, showing a positive correlation between Status and Income\footnote{Additional plots for all variables are available at \url{https://github.com/mimuc/labeler-bias}\label{note:second_github}}

\begin{figure*}[t]
    \centering
    \includegraphics[width=\textwidth]{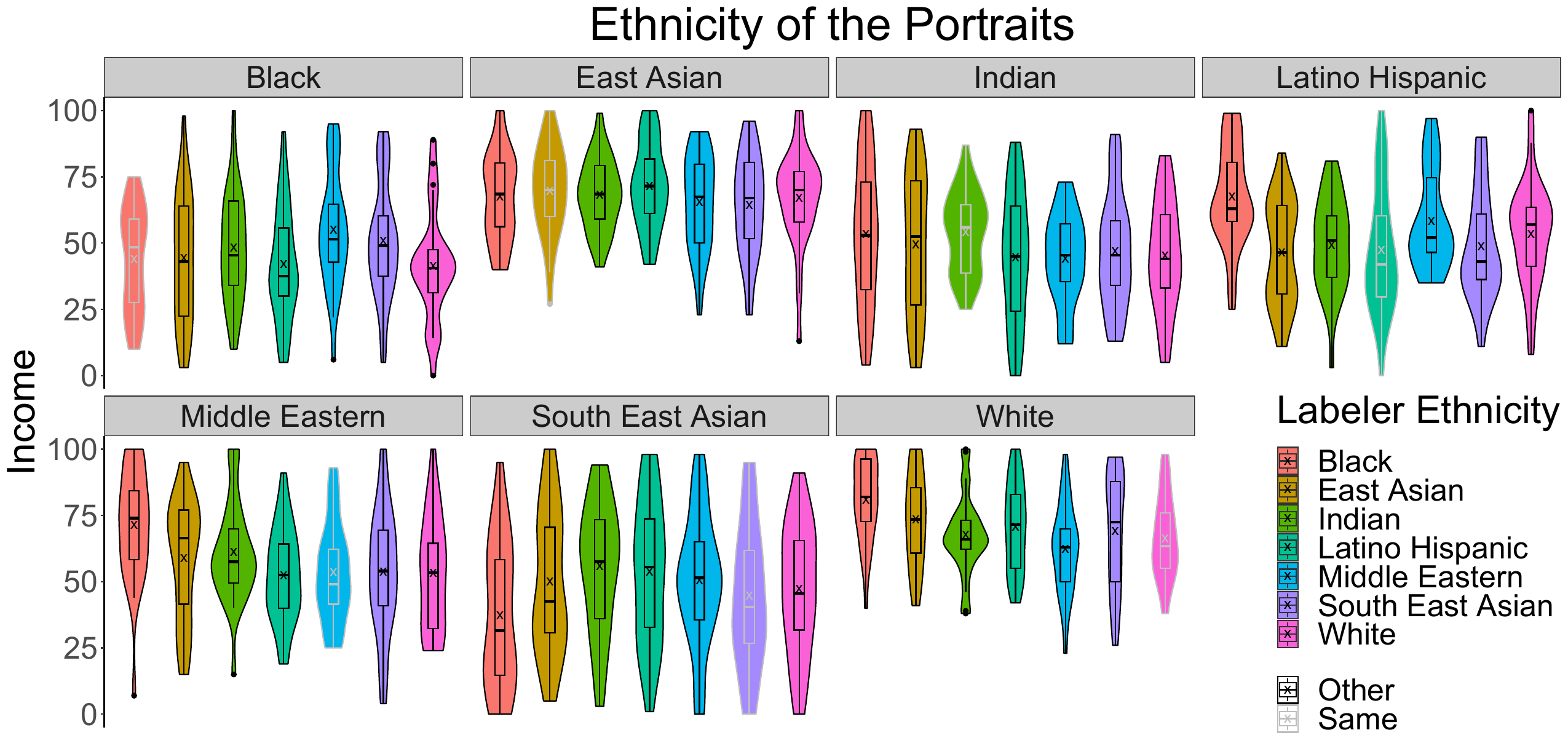}
    \caption{Estimated income as a function of \textsc{Labeler$_{Ethnicity}$} and  \textsc{Portrait$_{Ethnicity}$}. Grey borders indicate the cases where \textsc{Labeler$_{Ethnicity}$} and \textsc{Portrait$_{Ethnicity}$} match.}
    \label{fig:labeler_ethn_bar}
\end{figure*}

\subsection{The Impact of Demographics on Estimations (RQ2)}
Our second research question asks whether participant demographics impact their perceptions of stereotypes and their income estimations. \autoref{fig:labeler_ethn_bar} shows the income labels assigned to each \textsc{Portrait$_{Ethnicity}$} by each \textsc{Labeler$_{Ethnicity}$}\textsuperscript{\ref{note:second_github}}.

\begin{table*}[t]
	\caption{The two-way ANOVA results for the Income estimates and the four stereotype variables for \textsc{Labeler$_{Ethnicity}$} and \textsc{Portrait$_{Ethnicity}$}.}
	\label{tab:ethnicity_anova}
	\centering
	 \resizebox{\linewidth}{!}{
	\begin{tabular}{lccd{2.3}d{1.2}d{1.3}d{2.2}d{3.0}d{2.2}d{1.3}d{1.3}d{2.1}d{3.1}d{1.3}d{1.3}d{1.2}}
		\toprule 
&  \multicolumn{5}{c}{\textsc{Labeler}$_{ethnicity}$}  & \multicolumn{5}{c}{\textsc{Portrait}$_{ethnicity}$} &  \multicolumn{5}{c}{\textsc{L} $\times$ \textsc{P}}\\
\cmidrule(r){2-6}\cmidrule(lr){7-11}  \cmidrule(l){12-16} 
& \multicolumn{1}{c}{\textit{df}} & \multicolumn{1}{c}{\textit{dfs}} & \multicolumn{1}{c}{\textit{F}} & \multicolumn{1}{c}{\textit{p}} & \multicolumn{1}{c}{$\eta_{p}^{2}$} 
& \multicolumn{1}{c}{\textit{df}} & \multicolumn{1}{c}{\textit{dfs}} & \multicolumn{1}{c}{\textit{F}} & \multicolumn{1}{c}{\textit{p}}  & \multicolumn{1}{c}{$\eta_{p}^{2}$}
& \multicolumn{1}{c}{\textit{df}} & \multicolumn{1}{c}{\textit{dfs}} & \multicolumn{1}{c}{\textit{F}} & \multicolumn{1}{c}{\textit{p}}  & \multicolumn{1}{c}{$\eta_{p}^{2}$}\\
		\midrule
        Income      & 6 & 91   & 0.92 &                                .486  & .06 & 5.28 & 481  & 49.2 & \multicolumn{1}{B{.}{.}{1,3}}{<.001} & .35 & 31.7 & 481  & 2.39 & \multicolumn{1}{B{.}{.}{1,3}}{<.001} & .14 \\
        Warmth      & 6 & 91   & 1.2  &                                .313  & .07 & 4.92 & 448 &  3.9 & \multicolumn{1}{B{.}{.}{1,3}}{ .002} & .04 & 36.   & 546   & 1.35 &                                .089  & .08 \\
        Competence  & 6 & 91   & 0.71 &                                .642  & .04 & 6.   & 546   & 11.2 & \multicolumn{1}{B{.}{.}{1,3}}{<.001} & .11 & 36.   & 546   & 1.64 & \multicolumn{1}{B{.}{.}{1,3}}{ .012} & .1  \\
        Status      & 6 & 91   & 4.33 & \multicolumn{1}{B{.}{.}{1,3}}{<.001} & .22 & 6.   & 546   & 40.2 & \multicolumn{1}{B{.}{.}{1,3}}{<.001} & .31 & 36.   & 546   & 2.15 & \multicolumn{1}{B{.}{.}{1,3}}{<.001} & .12 \\
        Competition & 6 & 91   & 2.66 & \multicolumn{1}{B{.}{.}{1,3}}{ .02 } & .15 & 4.81 & 438 & 31.6 & \multicolumn{1}{B{.}{.}{1,3}}{<.001} & .26 & 28.9 & 438 & 1.84 & \multicolumn{1}{B{.}{.}{1,3}}{ .006} & .11 \\
		\bottomrule
	\end{tabular} 
	}
\end{table*}

We conducted an ANOVA using the interaction effect of \textsc{Labeler$_{Ethnicity}$} and \textsc{Portrait$_{Ethnicity}$}. The results, shown in \autoref{tab:ethnicity_anova}, reveal that the \textsc{Labeler$_{Ethnicity}$} significantly impacts status, the \textsc{Portrait$_{Ethnicity}$} significantly impacts all variables, and the interaction effect is significant for all variables except warmth. For an exemplary chart that shows income as a function of \textsc{Labeler$_{Ethnicity}$} and \textsc{Portrait$_{Ethnicity}$}, see \autoref{fig:labeler_ethn_bar}.

We also performed an ANOVA to investigate the impact of \textsc{Sex}. We found no significant main- or interaction-effects on income  (all $p >.05$), competence  (all $p > .05$),  status  (all $p > .05$) or competition  (all $p > .05$) estimates. Note however, that in the ANOVA on warmth estimates, we found a main effect of \textsc{Portrait$_{Sex}$}, $F(1, 96) = 10.06$, $p = .002$, $\eta_{p}^{2} = .09$. None of the other ANOVA-terms were significant (all $p > .05$).

\section{Discussion}
In this work, we set out to answer our two research questions. In the following, we discuss how our results address the research questions, the consequences for machine learning practice, and the limitations of our study.

\subsection{Labelers Exhibit Bias}
Our results provide evidence that labeler bias exists in two ways. First, our findings indicate that \textbf{labelers hold stereotypes about the people depicted in the portraits} they are tasked with annotating (see \autoref{fig:scm_portraits}) and that these stereotypes have an impact on their subsequent income labels (see \autoref{tab:pearson}). Second, \autoref{tab:ethnicity_anova} demonstrates that \textsc{Labeler$_{Ethnicity}$} and \textsc{Portrait$_{Ethnicity}$} have a significant impact on perceptions of stereotypes and income labels\footnote{Note that we did not test all comparisons post hoc as this procedure would not be informative to our study for two main reasons: First, one would need to apply very conservative $\alpha$-level corrections in order to avoid Type-II error inflation. Second, we were not interested in characterizing the specific bias of a certain group but rather intend to show that variation within the estimates can be explained by considering the interaction of labeler demographics and portrait demographics.}. Our results show that \textsc{Labeler$_{Ethnicity}$} significantly impacts status, \textsc{Portrait$_{Ethnicity}$} significantly impacts all variables, and the interaction effect is significant for all variables except warmth.

We found that income estimates were correlated with stereotype perceptions (\textbf{RQ1}) and that both stereotype perceptions and income estimates were impacted by \textsc{Labeler$_{Ethnicity}$} and the \textsc{Portrait$_{Ethnicity}$} (\textbf{RQ2}). Interestingly, \textsc{Sex} only had a significant effect on warmth for this task. This is in line with psychological research showing how stereotype judgments are made across cultures~\cite{cuddy2009stereotype,lin2005stereotype} and, in particular, how gender stereotypes influence perception~\cite{cuddy2008warmth}. In this domain, where labelers are annotating portraits of humans, we can conclude that labeler bias exists, depends on labeler demographics, and can be explained using stereotype content~\cite{fiske1991social}. These results are in line with recent findings in CSCW by Goyal et al. \cite{goyal_is_2022} demonstrating that toxicity labels for online content are influenced by labeler self-identification. 

\subsection{Implications for Machine Learning}
Our results shed some light on the impact of the human-aspect of machine learning. We have shown that labels vary with labeler demographics for annotation tasks involving portraits. This is important because prior work has shown that the majority of workers on MTurk are from the United States and India~\cite{difallah2018demographics,ipeirotis2010demographics,ross2010who}, and Levay et al.~\cite{levay2016demographic} found that over 70\% of MTurkers identify as white. Consequently, the status quo of gathering labels on crowdsourcing platforms without considering \textit{who} is doing the labeling should be reconsidered, as it will almost certainly lead to a non-balanced labeler pool and subsequently biased labels. However, as shown in \autoref{fig:scm_portraits} and \autoref{tab:ethnicity_anova}, stereotypes occur in labeling tasks involving images of people independent of labeler demographics. Recruiting labelers from a wide variety of populations should result in labels that are biased in a way that is consistent with societal biases, but the bias will still exist. Therefore, it remains an open research question as to whether it is possible to combat labeler bias through recruiting or to correct it post hoc. Post hoc methods have been proposed for bias in machine learning labels (e.g., \cite{jiang_identifying_2020}), but typically require knowledge of ground truth. Ground truths for social issues are complicated by the fact that representation in data is connected to the unequal distribution of power in society~\cite{dignazio_data_2020}. On such social issues, where ground truths may be fuzzy or non-existent, and society may be biased against particular groups, there is space for crucial future research to develop solutions. Past work by Miceli et al. \cite{miceli_studying_2022} suggests expanding data documentation and incorporating social contexts, which could be an important step toward ensuring fair, transparent data pipelines.

\subsection{Limitations \& Future Work}\label{section:limitations}
The most notable limitation of our study is that we have only explored one specific labeling task, namely annotating secondary characteristics of faces. This labeling scenario was chosen because it was likely to highlight the impact of stereotypes while still being rooted in a realistic scenario, such as making hiring decisions, which have been shown to be impacted by stereotypes~\cite{gonzalez_role_2019}. However, further work is required to understand how stereotypes and labeler demographics impact more abstract labeling tasks, such as image segmentation or product labeling. There are many high-stakes scenarios relevant to specific tasks across the field of machine learning that may be impacted by this phenomenon. Since we have now shown that there is a bias in this scenario, we call on future work to explore additional scenarios that are relevant to specific tasks in machine learning. For example, labeler bias may be relevant in detecting and classifying objects in autonomous driving tasks. Quantifying how these results generalize to other tasks is crucial to understanding when labeler demographics and their interactions with content must be accounted for.

Another limitation derives from the methodology used to create the FairFace dataset. The ethnicity, sex, and age labels in the dataset were created using a majority vote method on Amazon Mechanical Turk. The labels, therefore, are not necessarily a ground truth. Similarly, there is no ground truth for the income estimates since this information was not included in the FairFace dataset. Despite these limitations, this is the best available dataset we could locate with a balanced representation and labeled demographics and was, therefore, the best option available for this work. Future studies should investigate the magnitude of bias as a difference from the ground truth. Resolving this issue is not trivial, as it likely requires that a new database of images be generated with associated income levels provided by the image subjects, but it would be a worthy endeavor to further investigate and characterize this information.

\section{Conclusion}
In this paper, we investigated the existence and impact of labeler bias. We recruited 98 participants to engage in an online survey where we asked them to estimate the income and rate the perceived warmth, competence, status, and competition of people from multiple ethnicities and sexes portrayed in a series of images. We selected the portraits from the FairFace dataset using predefined exclusion criteria to create a balanced set of portraits. We found a significant relationship between income estimates and perceptions of competence, status, and competition. Additionally, the results indicate that the main- and interaction-effects of \textsc{Labeler$_{Ethnicity}$} and \textsc{Portrait$_{Ethnicity}$} significantly affect estimations, while \textsc{Labeler$_{Sex}$} and \textsc{Portrait$_{Sex}$} did not significantly impact the results. This insight poses a major challenge for AI applications, as it implies that datasets annotated by a non-diverse set of labelers are likely to carry stereotypes. Thus, we recommend that further research develops guidelines for responsible dataset generation and that researchers and practitioners reconsider the status quo for data labeling.



\bibliographystyle{vancouver}
\bibliography{bibliography}

\begin{thebibliography}{10}

\bibitem{horwitz2021facebook}
Hagey K, Horwitz J.
\newblock Facebook Tried to Make Its Platform a Healthier Place. It Got Angrier Instead.
\newblock Wall Street Journal. 2021.
\newblock Available from: \url{https://www.wsj.com/articles/facebook-algorithm-change-zuckerberg-11631654215}.

\bibitem{eickhoff2018cognitive}
Eickhoff C.
\newblock Cognitive Biases in Crowdsourcing.
\newblock In: Proceedings of the Eleventh ACM International Conference on Web Search and Data Mining. WSDM '18. New York, NY, USA: Association for Computing Machinery; 2018. p. 162–170.
\newblock \href {http://dx.doi.org/10.1145/3159652.3159654} {doi:10.1145/3159652.3159654}.

\bibitem{miceli_between_2020}
Miceli M, Schuessler M, Yang T.
\newblock Between Subjectivity and Imposition: Power Dynamics in Data Annotation for Computer Vision.
\newblock Proc ACM Hum-Comput Interact. 2020 oct;4(CSCW2).
\newblock \href {http://dx.doi.org/10.1145/3415186} {doi:10.1145/3415186}.

\bibitem{miceli_studying_2022}
Miceli M, Posada J, Yang T.
\newblock Studying Up Machine Learning Data: Why Talk About Bias When We Mean Power?
\newblock Proceedings of the ACM on Human-Computer Interaction. 2022 Jan;6(GROUP):34:1-34:14.
\newblock \href {http://dx.doi.org/10.1145/3492853} {doi:10.1145/3492853}.

\bibitem{goyal_is_2022}
Goyal N, Kivlichan ID, Rosen R, Vasserman L.
\newblock Is Your Toxicity My Toxicity? Exploring the Impact of Rater Identity on Toxicity Annotation.
\newblock Proc ACM Hum-Comput Interact. 2022 nov;6(CSCW2).
\newblock \href {http://dx.doi.org/10.1145/3555088} {doi:10.1145/3555088}.

\bibitem{sap_annotators_2022}
Sap M, Swayamdipta S, Vianna L, Zhou X, Choi Y, Smith NA.
\newblock Annotators with Attitudes: How Annotator Beliefs And Identities Bias Toxic Language Detection; 2022.
\newblock \href {http://dx.doi.org/10.48550/arXiv.2111.07997} {doi:10.48550/arXiv.2111.07997}.

\bibitem{bender2021dangers}
Bender EM, Gebru T, McMillan-Major A, Shmitchell S.
\newblock On the Dangers of Stochastic Parrots: Can Language Models Be Too Big?
\newblock In: Proceedings of the 2021 ACM Conference on Fairness, Accountability, and Transparency. FAccT '21. New York, NY, USA: Association for Computing Machinery; 2021. p. 610–623.
\newblock \href {http://dx.doi.org/10.1145/3442188.3445922} {doi:10.1145/3442188.3445922}.

\bibitem{jiang_identifying_2020}
Jiang H, Nachum O.
\newblock Identifying and Correcting Label Bias in Machine Learning.
\newblock In: Proceedings of the Twenty Third International Conference on Artificial Intelligence and Statistics. PMLR; 2020. p. 702-12.
\newblock Available from: \url{https://proceedings.mlr.press/v108/jiang20a.html}.

\bibitem{wauthier_bayesian_2011}
Wauthier FL, Jordan M.
\newblock Bayesian Bias Mitigation for Crowdsourcing.
\newblock In: Advances in Neural Information Processing Systems. vol.~24. Curran Associates, Inc.; 2011. p. 1-9.
\newblock Available from: \url{https://proceedings.neurips.cc/paper/2011/hash/0768281a05da9f27df178b5c39a51263-Abstract.html}.

\bibitem{cohn_modelling_2013}
Cohn T, Specia L.
\newblock Modelling Annotator Bias with Multi-task Gaussian Processes: An Application to Machine Translation Quality Estimation.
\newblock In: Proceedings of the 51st Annual Meeting of the Association for Computational Linguistics (Volume 1: Long Papers). Sofia, Bulgaria: Association for Computational Linguistics; 2013. p. 32-42.
\newblock Available from: \url{https://aclanthology.org/P13-1004}.

\bibitem{fiske2018stereotype}
Fiske ST.
\newblock Stereotype Content: Warmth and Competence Endure.
\newblock Current Directions in Psychological Science. 2018;27(2).
\newblock \href {http://dx.doi.org/10.1177/0963721417738825} {doi:10.1177/0963721417738825}.

\bibitem{schwind2019understanding}
Schwind V, Deierlein N, Poguntke R, Henze N.
\newblock Understanding the Social Acceptability of Mobile Devices using the Stereotype Content Model.
\newblock In: Proceedings of the 2019 CHI Conference on Human Factors in Computing Systems. New York, NY, USA: Association for Computing Machinery; 2019. p. 1–12.
\newblock \href {http://dx.doi.org/10.1145/3290605.3300591} {doi:10.1145/3290605.3300591}.

\bibitem{marsden_stereotypes_2016}
Marsden N, Haag M.
\newblock Stereotypes and Politics: Reflections on Personas.
\newblock In: Proceedings of the 2016 CHI Conference on Human Factors in Computing Systems. CHI '16. New York, NY, USA: Association for Computing Machinery; 2016. p. 4017-31.
\newblock \href {http://dx.doi.org/10.1145/2858036.2858151} {doi:10.1145/2858036.2858151}.

\bibitem{karkkainen2021fairface}
Karkkainen K, Joo J.
\newblock FairFace: Face Attribute Dataset for Balanced Race, Gender, and Age for Bias Measurement and Mitigation.
\newblock In: Winter Conference on Applications of Computer Vision (WACV). Waikoloa, HI, USA: IEEE; 2021. p. 1547-57.
\newblock \href {http://dx.doi.org/10.1109/WACV48630.2021.00159} {doi:10.1109/WACV48630.2021.00159}.

\bibitem{hagendorff2020ethics}
Hagendorff T.
\newblock The Ethics of AI Ethics: An Evaluation of Guidelines.
\newblock Minds \& Machines. 2020;30(1).
\newblock \href {http://dx.doi.org/10.1007/s11023-020-09517-8} {doi:10.1007/s11023-020-09517-8}.

\bibitem{tsamados2021ethics}
Tsamados A, Aggarwal N, Cowls J, Morley J, Roberts H, Taddeo M, et~al.
\newblock The ethics of algorithms: key problems and solutions.
\newblock AI \& Soc. 2021.
\newblock \href {http://dx.doi.org/10.1007/s00146-021-01154-8} {doi:10.1007/s00146-021-01154-8}.

\bibitem{muller2021ethics}
Mueller VC.
\newblock Ethics of Artificial Intelligence and Robotics.
\newblock The Stanford Encyclopedia of Philosophy. 2021.
\newblock Available from: \url{https://plato.stanford.edu/archives/sum2021/entries/ethics-ai/}.

\bibitem{schwartz2019untold}
Schwartz O.
\newblock Untold history of AI: Algorithmic bias was born in the 1980s.
\newblock IEEE Spectrum. 2019.
\newblock Available from: \url{https://spectrum.ieee.org/untold-history-of-ai-the-birth-of-machine-bias}.

\bibitem{lee2019algorithmic}
Lee NT, Resnick P, Barton G.
\newblock Algorithmic bias detection and mitigation: Best practices and policies to reduce consumer harms.
\newblock Brookings Institute: Washington, DC, USA. 2019.
\newblock Available from: \url{https://policycommons.net/artifacts/4141276/algorithmic-bias-detection-and-mitigation/4949849/}.

\bibitem{kodiyan2019overview}
Kodiyan AA.
\newblock An overview of ethical issues in using AI systems in hiring with a case study of Amazon’s AI based hiring tool; 2019.

\bibitem{zarsky2015understanding}
Zarsky T.
\newblock Understanding Discrimination in the Scored Society.
\newblock Rochester, NY: Social Science Research Network; 2015. ID 2550248.
\newblock Available from: \url{https://papers.ssrn.com/abstract=2550248}.

\bibitem{angwin2016machine}
Angwin J, Larson J, Mattu S, Kirchner L.
\newblock Machine bias.
\newblock Ethics of Data and Analytics. 2016:254-64.
\newblock \href {http://dx.doi.org/10.1201/9781003278290-37} {doi:10.1201/9781003278290-37}.

\bibitem{barera2020mind}
Barera M. Mind the Gap: Addressing Structural Equity and Inclusion on Wikipedia. University of Texas at Arlington: University of Texas at Arlington; 2020.
\newblock Available from: \url{http://hdl.handle.net/10106/29572}.

\bibitem{silva_effects_2015}
Silva FAMd, Pedrini H.
\newblock Effects of cultural characteristics on building an emotion classifier through facial expression analysis.
\newblock Journal of Electronic Imaging. 2015 Mar;24(2):023015.
\newblock \href {http://dx.doi.org/10.1117/1.JEI.24.2.023015} {doi:10.1117/1.JEI.24.2.023015}.

\bibitem{scheuerman_how_2020}
Scheuerman MK, Wade K, Lustig C, Brubaker JR.
\newblock How We've Taught Algorithms to See Identity: Constructing Race and Gender in Image Databases for Facial Analysis.
\newblock Proceedings of the ACM on Human-Computer Interaction. 2020 May;4(CSCW1):58:1-58:35.
\newblock \href {http://dx.doi.org/10.1145/3392866} {doi:10.1145/3392866}.

\bibitem{wall2018four}
Wall E, Blaha LM, Paul CL, Cook K, Endert A.
\newblock Four Perspectives on Human Bias in Visual Analytics.
\newblock In: Cognitive Biases in Visualizations. Cham: Springer International Publishing; 2018. p. 29-42.
\newblock \href {http://dx.doi.org/10.1007/978-3-319-95831-6\_3} {doi:10.1007/978-3-319-95831-6\_3}.

\bibitem{hutson2018debiasing}
Hutson J, Taft JG, Barocas S, Levy K.
\newblock Debiasing Desire: Addressing Bias \& Discrimination on Intimate Platforms.
\newblock Proc ACM Hum-Comput Interact. 2018;2(CSCW).
\newblock \href {http://dx.doi.org/10.1145/3274342} {doi:10.1145/3274342}.

\bibitem{wick_unlocking_2019}
Wick M, Panda S, Tristan JB.
\newblock Unlocking Fairness: a Trade-off Revisited.
\newblock Advances in neural information processing systems. 2019;32.
\newblock Available from: \url{https://proceedings.neurips.cc/paper/2019/file/373e4c5d8edfa8b74fd4b6791d0cf6dc-Paper.pdf}.

\bibitem{hube_understanding_2019}
Hube C, Fetahu B, Gadiraju U.
\newblock Understanding and Mitigating Worker Biases in the Crowdsourced Collection of Subjective Judgments.
\newblock In: Proceedings of the 2019 CHI Conference on Human Factors in Computing Systems. CHI '19. New York, NY, USA: Association for Computing Machinery; 2019. p. 1-12.
\newblock \href {http://dx.doi.org/10.1145/3290605.3300637} {doi:10.1145/3290605.3300637}.

\bibitem{barbosa2019rehumanized}
Barbosa NaM, Chen M.
\newblock Rehumanized Crowdsourcing: A Labeling Framework Addressing Bias and Ethics in Machine Learning.
\newblock In: Proceedings of the 2019 CHI Conference on Human Factors in Computing Systems. CHI '19. New York, NY, USA: Association for Computing Machinery; 2019. p. 1–12.
\newblock \href {http://dx.doi.org/10.1145/3290605.3300773} {doi:10.1145/3290605.3300773}.

\bibitem{geva_are_2019}
Geva M, Goldberg Y, Berant J.
\newblock Are We Modeling the Task or the Annotator? An Investigation of Annotator Bias in Natural Language Understanding Datasets; 2019.
\newblock \href {http://dx.doi.org/10.48550/arXiv.1908.07898} {doi:10.48550/arXiv.1908.07898}.

\bibitem{parmar_dont_2022}
Parmar M, Mishra S, Geva M, Baral C.
\newblock Don't Blame the Annotator: Bias Already Starts in the Annotation Instructions; 2022.
\newblock \href {http://dx.doi.org/10.48550/arXiv.2205.00415} {doi:10.48550/arXiv.2205.00415}.

\bibitem{miceli_data-production_2022}
Miceli M, Posada J.
\newblock The Data-Production Dispositif; 2022.
\newblock \href {http://dx.doi.org/10.48550/arXiv.2205.11963} {doi:10.48550/arXiv.2205.11963}.

\bibitem{engelmann_what_2022}
Engelmann S, Ullstein C, Papakyriakopoulos O, Grossklags J.
\newblock What People Think AI Should Infer From Faces.
\newblock In: 2022 ACM Conference on Fairness, Accountability, and Transparency. FAccT '22. New York, NY, USA: Association for Computing Machinery; 2022. p. 128-41.
\newblock \href {http://dx.doi.org/10.1145/3531146.3533080} {doi:10.1145/3531146.3533080}.

\bibitem{fiske2007universal}
Fiske ST, Cuddy AJC, Glick P.
\newblock Universal dimensions of social cognition: Warmth and competence.
\newblock Trends in cognitive sciences. 2007;11(2).
\newblock \href {http://dx.doi.org/10.1016/j.tics.2006.11.005} {doi:10.1016/j.tics.2006.11.005}.

\bibitem{fiske2002model}
Fiske ST, Cuddy AJC, Glick P, Xu J.
\newblock A model of (often mixed) stereotype content: Competence and warmth respectively follow from perceived status and competition.
\newblock Journal of Personality and Social Psychology. 2002;82(6).
\newblock \href {http://dx.doi.org/10.1037/0022-3514.82.6.878} {doi:10.1037/0022-3514.82.6.878}.

\bibitem{dovidio2010intergroup}
Dovidio JF, Gaertner SL.
\newblock Intergroup bias.
\newblock In: Handbook of social psychology, Vol. 2, 5th ed. Hoboken, NJ, US: John Wiley \& Sons, Inc.; 2010. p. 1084-121.
\newblock \href {http://dx.doi.org/10.1002/9780470561119.socpsy002029} {doi:10.1002/9780470561119.socpsy002029}.

\bibitem{durante2017poor}
Durante F, Tablante CB, Fiske ST.
\newblock Poor but Warm, Rich but Cold (and Competent): Social Classes in the Stereotype Content Model.
\newblock Journal of Social Issues. 2017;73(1).
\newblock \href {http://dx.doi.org/10.1111/josi.12208} {doi:10.1111/josi.12208}.

\bibitem{grigoryev2019mapping}
Grigoryev D, Fiske ST, Batkhina A.
\newblock Mapping Ethnic Stereotypes and Their Antecedents in Russia: The Stereotype Content Model.
\newblock Frontiers in Psychology. 2019;10.
\newblock \href {http://dx.doi.org/10.3389/fpsyg.2019.01643} {doi:10.3389/fpsyg.2019.01643}.

\bibitem{diamantopoulos2017explicit}
Diamantopoulos A, Florack A, Halkias G, Palcu J.
\newblock Explicit versus implicit country stereotypes as predictors of product preferences: Insights from the stereotype content model.
\newblock J Int Bus Stud. 2017;48(8).
\newblock \href {http://dx.doi.org/10.1057/s41267-017-0085-9} {doi:10.1057/s41267-017-0085-9}.

\bibitem{lin2005stereotype}
Lin MH, Kwan VSY, Cheung A, Fiske ST.
\newblock Stereotype Content Model Explains Prejudice for an Envied Outgroup: Scale of Anti-Asian American Stereotypes.
\newblock Pers Soc Psychol Bull. 2005;31(1).
\newblock \href {http://dx.doi.org/10.1177/0146167204271320} {doi:10.1177/0146167204271320}.

\bibitem{cuddy2008warmth}
Cuddy AJC, Fiske ST, Glick P.
\newblock Warmth and Competence as Universal Dimensions of Social Perception: The Stereotype Content Model and the BIAS Map.
\newblock Advances in experimental social psychology. 2008;40:61-149.
\newblock \href {http://dx.doi.org/10.1016/S0065-2601(07)00002-0} {doi:10.1016/S0065-2601(07)00002-0}.

\bibitem{fiske2015intergroup}
Fiske ST.
\newblock Intergroup Biases: A Focus on Stereotype Content.
\newblock Curr Opin Behav Sci. 2015;3.
\newblock \href {http://dx.doi.org/10.1016/j.cobeha.2015.01.010} {doi:10.1016/j.cobeha.2015.01.010}.

\bibitem{cuddy2009stereotype}
Cuddy AJC, Fiske ST, Kwan VSY, Glick P, Demoulin S, Leyens JP, et~al.
\newblock Stereotype content model across cultures: Towards universal similarities and some differences.
\newblock British Journal of Social Psychology. 2009;48(1).
\newblock \href {http://dx.doi.org/10.1348/014466608X314935} {doi:10.1348/014466608X314935}.

\bibitem{nicolas2021comprehensive}
Nicolas G, Bai X, Fiske ST.
\newblock Comprehensive stereotype content dictionaries using a semi-automated method.
\newblock European Journal of Social Psychology. 2021;51(1).
\newblock \href {http://dx.doi.org/10.1002/ejsp.2724} {doi:10.1002/ejsp.2724}.

\bibitem{kroon2021guilty}
Kroon AC, Trilling D, Raats T.
\newblock Guilty by Association: Using Word Embeddings to Measure Ethnic Stereotypes in News Coverage.
\newblock Journalism \& Mass Communication Quarterly. 2021;98(2).
\newblock \href {http://dx.doi.org/10.1177/1077699020932304} {doi:10.1177/1077699020932304}.

\bibitem{fraser2021understanding}
Fraser KC, Nejadgholi I, Kiritchenko S.
\newblock Understanding and Countering Stereotypes: A Computational Approach to the Stereotype Content Model; 2021.
\newblock \href {http://dx.doi.org/10.48550/arXiv.2106.02596} {doi:10.48550/arXiv.2106.02596}.

\bibitem{mckee2021understanding}
McKee K, Bai X, Fiske S.
\newblock Understanding Human Impressions of Artificial Intelligence; 2021.
\newblock \href {http://dx.doi.org/10.31234/osf.io/5ursp} {doi:10.31234/osf.io/5ursp}.

\bibitem{phinney1996when}
Phinney JS.
\newblock When we talk about American ethnic groups, what do we mean?
\newblock American Psychologist. 1996;51(9).
\newblock \href {http://dx.doi.org/10.1037/0003-066X.51.9.918} {doi:10.1037/0003-066X.51.9.918}.

\bibitem{zhang2016learning}
Zhang J, Wu X, Sheng V.
\newblock Learning from crowdsourced labeled data: a survey.
\newblock Artificial Intelligence Review. 2016;46.
\newblock \href {http://dx.doi.org/10.1007/s10462-016-9491-9} {doi:10.1007/s10462-016-9491-9}.

\bibitem{lugaresi2019mediapipe}
Lugaresi C, Tang J, Nash H, McClanahan C, Uboweja E, Hays M, et~al.
\newblock MediaPipe: A Framework for Building Perception Pipelines; 2019.
\newblock \href {http://dx.doi.org/10.48550/arXiv.1906.08172} {doi:10.48550/arXiv.1906.08172}.

\bibitem{abbey2017attention}
Abbey JD, Meloy MG.
\newblock Attention by design: Using attention checks to detect inattentive respondents and improve data quality.
\newblock Journal of Operations Management. 2017;53-56.
\newblock \href {http://dx.doi.org/10.1016/j.jom.2017.06.001} {doi:10.1016/j.jom.2017.06.001}.

\bibitem{greenhouse1959methods}
Greenhouse SW, Geisser S.
\newblock On methods in the analysis of profile data.
\newblock Psychometrika. 1959;24(2).
\newblock \href {http://dx.doi.org/10.1007/BF02289823} {doi:10.1007/BF02289823}.

\bibitem{fiske1991social}
Fiske ST, Taylor SE.
\newblock Social Cognition (2nd ed).
\newblock New York, NY, England: Mcgraw-Hill Book Company; 1991.
\newblock Available from: \url{https://psycnet.apa.org/record/1991-97723-000}.

\bibitem{difallah2018demographics}
Difallah D, Filatova E, Ipeirotis P.
\newblock Demographics and Dynamics of Mechanical Turk Workers.
\newblock In: Proceedings of the Eleventh ACM International Conference on Web Search and Data Mining. WSDM '18. New York, NY, USA: Association for Computing Machinery; 2018. p. 135–143.
\newblock \href {http://dx.doi.org/10.1145/3159652.3159661} {doi:10.1145/3159652.3159661}.

\bibitem{ipeirotis2010demographics}
Ipeirotis PG.
\newblock Demographics of Mechanical Turk.
\newblock Rochester, NY; 2010. 1585030.
\newblock Available from: \url{https://papers.ssrn.com/abstract=1585030}.

\bibitem{ross2010who}
Ross J, Irani L, Silberman MS, Zaldivar A, Tomlinson B.
\newblock Who are the crowdworkers? shifting demographics in mechanical turk.
\newblock In: CHI '10 Extended Abstracts on Human Factors in Computing Systems. CHI EA '10. New York, NY, USA: Association for Computing Machinery; 2010. p. 2863–2872.
\newblock \href {http://dx.doi.org/10.1145/1753846.1753873} {doi:10.1145/1753846.1753873}.

\bibitem{levay2016demographic}
Levay KE, Freese J, Druckman JN.
\newblock The Demographic and Political Composition of Mechanical Turk Samples.
\newblock SAGE Open. 2016;6(1).
\newblock \href {http://dx.doi.org/10.1177/2158244016636433} {doi:10.1177/2158244016636433}.

\bibitem{dignazio_data_2020}
D'Ignazio C, Klein LF.
\newblock Data Feminism.
\newblock MIT Press; 2020.
\newblock Available from: \url{https://mitpress.mit.edu/9780262547185/data-feminism/}.

\bibitem{gonzalez_role_2019}
González MJ, Cortina C, Rodríguez J.
\newblock The Role of Gender Stereotypes in Hiring: A Field Experiment.
\newblock European Sociological Review. 2019;35(2).
\newblock \href {http://dx.doi.org/10.1093/esr/jcy055} {doi:10.1093/esr/jcy055}.

\end{thebibliography}

\end{document}